\documentclass{article}

\PassOptionsToPackage{numbers,compress}{natbib}

\usepackage[preprint]{neurips_2024}

\usepackage[utf8]{inputenc} 
\usepackage[T1]{fontenc}    
\usepackage{hyperref}       
\usepackage{url}            
\usepackage{booktabs}       
\usepackage{amsfonts}       
\usepackage{nicefrac}       
\usepackage{microtype}      
\usepackage{xcolor}         
\usepackage{ulem}           

\usepackage{graphicx}
\usepackage{tabularx}
\usepackage{multirow}
\usepackage{subcaption}
\usepackage{pgfplots}
\usepackage{caption}
\pgfplotsset{compat=1.17}
\usepackage{makecell}
\usepackage{acronym}
\usepackage{amsmath}
\usepackage{mathrsfs}
\usepackage{float}

\title{Exploring Graph-Transformer Out-of-Distribution Generalization Abilities}

\author{%
  Itay Niv \\ 
  Faculty of Engineering \\
  Tel-Aviv University \\
  \texttt{itayniv@mail.tau.ac.il} \\ 
  \And
  Neta Rabin \\
  Faculty of Engineering \\
  Tel-Aviv University \\
  \texttt{netara@tauex.tau.ac.il} \\
}

\begin{document}
\acrodef{GNNs}{Graph neural networks}
\acrodef{DG}{domain generalization}
\acrodef{MPNNs}{message passing neural networks}
\acrodef{GT}{graph-transformer}
\acrodef{MHA}{multi head attention}
\acrodef{ID}{in-distribution}
\acrodef{OOD}{out-of-distribution}
\acrodef{i.i.d}{independent and identically distributed}
\acrodef{ERM}{empirical risk minimization}
\acrodef{vGIN}{virtual-GIN}
\acrodef{MMD}{maximum mean discrepancy}
\acrodef{t-SNE}{t-distributed stochastic neighbor embedding}
\acrodef{GPS}{GraphGPS}
\acrodef{SOTA}{state-of-the-art}
\acrodef{RW}{random-walk}
\acrodef{GOOD}{Graph Out-of-Distribution}
\acrodef{NLP}{natural-language-processing}

\maketitle

\begin{abstract}
Deep learning on graphs has shown remarkable success across numerous applications, including social networks, bio-physics, traffic networks, and recommendation systems. Regardless of their successes, current methods frequently depend on the assumption that training and testing data share the same distribution, a condition rarely met in real-world scenarios. While graph-transformer (GT) backbones have recently outperformed traditional message-passing neural networks (MPNNs) in multiple in-distribution (ID) benchmarks, their effectiveness under distribution shifts remains largely unexplored.

In this work, we address the challenge of out-of-distribution (OOD) generalization for graph neural networks, with a special focus on the impact of backbone architecture. We systematically evaluate GT and hybrid backbones in OOD settings and compare them to MPNNs. To do so, we adapt several leading domain generalization (DG) algorithms to work with GTs and assess their performance on a benchmark designed to test a variety of distribution shifts. Our results reveal that GT and hybrid GT-MPNN backbones demonstrate stronger generalization ability compared to MPNNs, even without specialized DG algorithms (on four out of six benchmarks). 

Additionally, we propose a novel post-training analysis approach that compares the clustering structure of the entire ID and OOD test datasets, specifically examining domain alignment and class separation. Highlighting its model-agnostic design, the method yielded valuable insights into both GT and MPNN backbones and appears well suited for broader DG applications beyond graph learning, offering a deeper perspective on generalization abilities that goes beyond standard accuracy metrics. Together, our findings highlight the promise of graph-transformers for robust, real-world graph learning and set a new direction for future research in OOD generalization.

\end{abstract}




\section{Introduction}\label{sec:intro}

Deep learning on graphs is both promising and rapidly evolving, with numerous challenges as well as exciting prospects. \ac{GNNs} have demonstrated strong results in various tasks \cite{wu2022graph}, including graph classification. However, their irregular topologies, unlike the sequence or grid structures found in text, audio, and images, make mathematical operations like convolution and pooling challenging to apply. To process graph data effectively, the first critical challenge is to learn graph data representation, which makes it easier to extract useful information, preserve graph structures, and discriminate information when building classifiers.
Pivotal developments in the field of \ac{GNNs} involve the creation of architectures, including Graph Convolution Network (GCN) \cite{kipf2016semi}, Graph Isomorphism Network (GIN) \cite{xu2018powerful}, and more.
They are primarily expressive within the bounds of the 1-Weisfeiler-Lehman (1-WL) isomorphism test because of their \ac{MPNNs} architecture \cite{xu2018powerful}. Recent studies have explored ways to improve the expressivity, for example, by enhancing features like positional encoding or structural information \cite{zhang2024expressive}.

A key challenge nowadays is \textit{\ac{OOD} generalization} \cite{li2022out}, which is essential for ensuring robustness and reliability in real-world settings. In particular, this work focuses on the \textit{\ac{DG}} problem, where models need to perform well on test domains they did not see during training. These domains correspond to distinct, yet potentially overlapping, distributions of data sharing the same prediction task. Most \ac{GNNs} rely on the \ac{i.i.d} assumption, where training and test data share the same distribution. Nevertheless, even with specialized \ac{DG} algorithms, they often experience significant performance drops under distribution shifts \cite{good_leaderboard}. Recent studies attribute this to the complexity of distribution shifts and frequent violations of key assumptions in practical settings, which also complicate the reliable evaluation of model generalization \cite{ma2024survey}.

Robust evaluation under OOD conditions is thus crucial for deploying models in high-stakes applications.
However, this evaluation is challenging, as traditional \ac{i.i.d} benchmarks are inadequate and require dedicated datasets and metrics to reflect realistic distribution shifts. The \textit{\ac{GOOD} benchmark} \cite{gui2022good} addresses this by providing diverse data splits with various distribution shifts. Still, it reports results solely on a single backbone, \ac{vGIN} \cite{xu2018powerful, gilmer2017neural}, with DG algorithms trained on top to enhance OOD robustness.  While a few recent works have begun to explore the impact of different backbones on OOD generalization \cite{lu2024graph, guo2024investigating}, they remain limited, focusing only on MPNN variants or node-level tasks, often excluding strong-performing baselines like GIN. Although our prior work was the first to investigate graph transformers in OOD settings, a recent concurrent and independent study, GOODFormer \cite{liao2025invariantgraphtransformeroutofdistribution}, proposed a DG algorithm for transformers that jointly optimizes invariant subgraph disentanglement and evolving subgraph encoding. Their focus was primarily on the DG algorithm rather than systematically evaluating the backbone itself,
which resulted in the transformer-based backbones not achieving their full performance potential on most benchmarks. These observations emphasize the need to systematically study how backbone architectures influence OOD performance independently of DG algorithms, and whether modern designs, such as transformer-based backbones, can enhance robustness to distribution shifts.

Motivated by the success of the classical transformer architecture, widely influential in representation learning for fields like \ac{NLP} \cite{brown2020language},
\ac{GT} models have been developed to incorporate global attention mechanisms and richer positional encodings. 
Notable examples include \ac{GPS} \cite{rampavsek2022recipe}, which is a modular framework for constructing networks by combining local message-passing and a global \ac{MHA} mechanism with a choice of various positional and structural encodings. SAN \cite{kreuzer2021rethinking} makes use of two attention mechanisms, one on the fully connected graph and one on the original edges of the input graph, while using Laplacian positional encodings for the nodes. Graphormer \cite{ying2021transformers} biases the attention weights using structural features in the form of centrality and spatial encodings. Exphormer \cite{shirzad2023exphormer} enhances \ac{GPS} by utilizing \ac{GNNs} original and expander graphs. These GT models have emerged as a popular alternative to traditional \ac{MPNNs} on standard i.i.d. benchmarks \cite{hoang2024survey} (Appendix \ref{sec:gt_mpnn_id_summary} summarizes prior work on this topic); however, their effectiveness under distribution shifts, critical for real-world deployment, remains largely unexplored.




In computer vision, transformers have shown robustness to certain distribution shifts compared to convolutional networks~\cite{zhang2022delving, li2022sparse},  motivating the exploration of similar benefits in graph learning. We hypothesize that transformers backbones offer key advantages over traditional \ac{MPNNs} for \ac{OOD} generalization in graph learning, based on both architectural and theoretical insight and empirical evidence from other domains. Their ability to model global context and long-range dependencies may help overcome limitations of \ac{MPNNs}, such as over-smoothing and constrained receptive fields. Despite this potential, the \ac{OOD} generalization capabilities of \ac{GT} remain underexplored. Moreover, hybrid architectures that combine local message-passing with global attention mechanisms may offer additional generalization gains, highlighting the need for systematic empirical evaluation of transformer-based graph backbones.

Besides the exploration and influence of the backbone of OOD performance, another key open question in the study of generalization abilities is how to evaluate and interpret OOD performance effectively. Most existing work relies primarily on accuracy-based metrics or intuitively inspecting features visualization \cite{sultana2022self}. In this work, we question whether accuracy alone is sufficient to evaluate OOD generalization and instead propose quantifying decoupled aspects that capture deeper model behavior. 
To this end, we employ post-hoc interpretability, analyzing a trained model without altering its architecture, by loading a model checkpoint and examining its internal representations across the full in-distribution (ID) and OOD test datasets.

When considering post-hoc interpretability, an important step is defining which model representations should be evaluated. For classification tasks, softmax outputs are often used, but they can be unreliable and overconfident for inputs that lie outside the training distribution \cite{gal2016uncertainty}. Higher-dimensional features from the penultimate layer are commonly used instead \cite{zhu2024croft, sun2022out}, as they tend to provide more robust uncertainty estimates and richer information for analysis. Beyond selecting the right representations, proper evaluation also demands dedicated metrics that capture decoupled aspects of generalization. Haotian  et al. \cite{ye2021towards} introduce the concepts of variation and informativeness as critical components for evaluating a model's ability to generalize to OOD data. Variation refers to the change in learned features across different distributions, while informativeness measures how well a feature distinguishes between classes. Misalignment in these learned features often signals a failure to generalize effectively to OOD inputs \cite{jegelka2022theory}. 

To move beyond the limitations of analyzing individual samples, we suggest a more global perspective on generalization evaluation. While some OOD detection methods effectively identify anomalies by comparing OOD sample features to ID training features alignment \cite{papernot2018deep}, they operate at the individual-level. In contrast, since OOD generalization aims to assess a model’s overall robustness, we argue that examining the global structure of both ID and OOD datasets provides deeper insight. To this end, we propose a cluster-wise analysis that compares the overall clustering structure of model representations across domains. First, we propose employing the squared \ac{MMD} \cite{gretton2012kernel} distance to measure alignment between ID and OOD features clusters, in a post-hoc manner. Li et al. have used MMD for OOD generalization in a different way, as a training loss to align source-domain features and learn domain-invariant representations. Second, we propose to use the Silhouette score \cite{rousseeuw1987silhouettes}, which evaluates how well each sample fits within its own class compared to other classes, capturing cluster cohesion and separation. Applied to concatenated ID and OOD features with class labels, it reveals how well the class structure is preserved across domains and whether OOD samples overlap or disrupt ID clusters. This approach enables to reveal patterns that are invisible to sample-wise comparisons, offering a more interpretable and comprehensive quantification of model robustness under distribution shift.

\subsection{Contributions}

\begin{itemize}
    \item We conduct the first systematic evaluation of GT backbones, including both pure GT and hybrid GT-MPNN (GPS) backbones, for OOD generalization in supervised graph classification. This includes adapting several DG algorithms to support GT backbones, and rigorously comparing them to an MPNN backbone (vGIN) across six diverse distribution shifts using the GOOD benchmark.
    \item We propose a novel post-hoc analysis framework that compares the clustering structures of the full ID and OOD test datasets in the model’s penultimate layer. By quantifying domain-alignment and class-separation using the proposed MMD and Silhouette metrics. It provides a decoupled, interpretable view of how learned representations behave under distribution shifts. Applied to GT and MPNN backbones, it revealed meaningful insights in model generalization abilities not captured by standard accuracy metrics. This model-agnostic approach is broadly applicable beyond graph learning.
    \item Our results show that, across both accuracy-based and the proposed metrics, the GPS backbone generally exhibits stronger generalization than MPNN and pure attention-based (MHA) models. When MHA performs best, its results are still similar to those of GPS.
\end{itemize}

\begin{figure}[t]
\centering
\begin{minipage}[t]{0.51\textwidth}
    \centering
    \begin{tikzpicture}[baseline]
    \begin{axis}[
        ybar,
        bar width=6pt,
        enlarge x limits=0.2,
        ylabel={OOD Accuracy},
        symbolic x coords={CMNIST-color, Motif-size, Motif-basis, SST2-length, HIV-size, HIV-scaffold},
        xtick=data,
        x tick label style={rotate=30, anchor=east, font=\tiny},
        ymin=0, ymax=100,
        legend style={
            at={(1,1.15)},
            font=\tiny,
            anchor=north east,legend columns=3
        },
        width=7cm,
        height=6cm
    ]
    \addplot+[style={fill=blue!50}, bar shift=-6pt] 
        coordinates {
            (CMNIST-color,29.845) 
            (Motif-size,52.235) 
            (Motif-basis,66.917) 
            (SST2-length,80.32)
            (HIV-size, 62.947) 
            (HIV-scaffold, 72.062)
        };
    
    \addplot+[style={fill=red!70}, bar shift=0pt] 
        coordinates {
            (CMNIST-color,23.4275) 
            (Motif-size,88.3425) 
            (Motif-basis,92.4825) 
            (SST2-length,78.235)
            (HIV-size, 53.937) 
            (HIV-scaffold, 60.612)
        };
    
    \addplot+[style={fill=green!50}, bar shift=6pt] 
        coordinates {
            (CMNIST-color,87.0275) 
            (Motif-size,84.2425) 
            (Motif-basis,92.1675) 
            (SST2-length,82.795)
            (HIV-size, 58.045) 
            (HIV-scaffold, 67.0475)
        };
    \legend{vGIN, MHA, GPS}
    \end{axis}
    \end{tikzpicture}
    \caption{OOD accuracy in \% of three backbones across six benchmarks. vGIN is equipped with \ac{DG} algorithms (which lead to the best OOD accuracy), while MHA and GPS are trained using standard ERM only. See Section \ref{sec:exp_details} for experimental details.}
    \label{fig:ood-bar}
\end{minipage}
\hfill
\begin{minipage}[t]{0.48\textwidth}
    \centering
    \includegraphics[width=\textwidth]{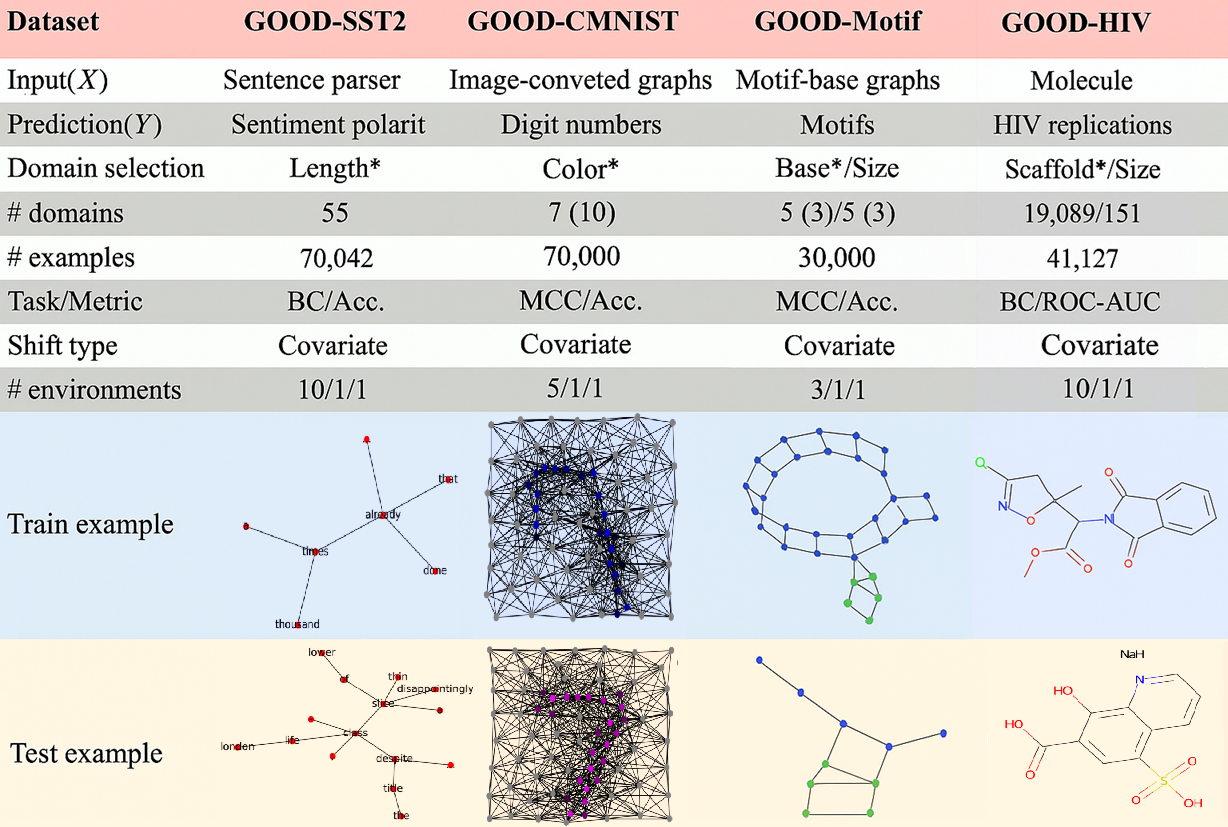}
    \caption{Overview of the GOOD benchmark graph classification tasks. Key properties and visual examples highlight distribution shifts across CMNIST, Motif, SST2, and HIV tasks. Figure adapted from S. Gui et al. \cite{gui2022good}.}
    \label{fig:datasets}
\end{minipage}
\end{figure}

\section{Problem definition}\label{sec:problem}

Generally, \ac{GNNs} can be decomposed into $f=\rho \circ h$ , where $h : G \rightarrow R^h$ is an econoder that learns a meaningful representation $h(G)$ for each graph $G$ to help predict the label $\hat{Y}_G = \rho(h(G))$ with a downstream classifier $\rho : \mathbb{R}^h \rightarrow Y$. 

This work focuses on OOD generalization in graph classification tasks. The problem can be defined as \ac{DG} that relies on incorporating domain-shift information.  Formally, we consider a dataset with a set of $|\mathcal{E}|$ domains $\mathcal{E} = \{e_1, . . . , e_{|E|}\}$. Samples $(G^e_i, Y^e_i) \in e$ from the same domain are considered to have been drawn independently from an identical distribution  $\mathbb{P}^e (G, Y)$. Let $\mathcal{E}_{all}$ be the domain set we want to generalize to, and the training and test domains $\mathcal{E}_{train}, \mathcal{E}_{test} \subset \mathcal{E}_{all}$. The goal of OOD generalization on graphs is to train a classifier $f^*$ with data from training domains $\mathcal{E}_{train}$ that generalizes well to all domains $\mathcal{E}_{all}$:

\begin{equation}
f^* = \underset{f}{\arg \min} \underset{G,Y \in \mathbb{P}^{\mathcal{E}_{all}}}{\mathbb{E}}[\mathcal{L}(f(G),Y)].
 \end{equation}

When $\mathbb{P}^{\mathcal{E}_{train}}(G,Y) = \mathbb{P}^{\mathcal{E}_{test}}(G,Y)$, the \ac{i.i.d} assumption holds, it becomes a standard learning problem.
In this setting, \ac{ERM} \cite{vapnik1999overview}, which directly minimizes the average training loss, can achieve good performance. Conversely, when  $\mathbb{P}^{\mathcal{E}_{train}}(G,Y) \neq \mathbb{P}^{\mathcal{E}_{test}}(G,Y)$, a distribution shift occurs, transforming it into an OOD generalization problem. In such instances, ERM may fail severely in the face of distribution shifts. 
To address the \ac{OOD} problem, three steps are required: distributional shift characterization, methodological development, and an \ac{OOD} generalization evaluation framework. 

\subsection{Distributional shift characterization}

Graphs from different domains might have, for example, different node features, sizes, or common structural patterns, which can make it harder for models to generalize. These distributional shifts can occur individually or in combination, where their distinct nature necessitates different approaches to tackling them \cite{gui2022good}. To this end, since no data from the test domains is available during training, DG algorithms must rely on assumptions about certain statistical invariances that hold across both training and test domains. These invariances guide the design of predictors that generalize beyond the training data. The specific type of invariance assumed and the method used to estimate it from the training domains differ across DG algorithms \cite{gulrajani2020search}.

In this work, we focus on \textit{covariate shift}, where the distribution of the input graph \( G \) changes between the training and test domains, while the conditional distribution of the target variable remains invariant. This setting is formally supported by the \textit{Independence of Cause and Mechanism (ICM)} principle~\cite{scholkopf2021toward}, which states that the distribution of the input (cause) is independent of the conditional mechanism generating the output (effect). As a result, domain shifts arise due to changes in \
$\mathbb{P}^{\mathcal{E}_{train}}(G) \neq \mathbb{P}^{\mathcal{E}_{test}}(G)$, 
while the predictive mechanism remains stable: \ 
$\mathbb{P}^{\mathcal{E}_{train}}(Y|G) = \mathbb{P}^{\mathcal{E}_{test}}(Y|G)$.

Another common assumption is that shifts between training and test distributions are partially reflected in the variation among training domains.  Specifically, if a feature is predictive and approximately invariant across training domains \(\mathcal{E}_{\text{train}}\), it will remain so in the broader environment \(\mathcal{E}_{\text{all}}\)~\cite{ye2021towards}. This allows models to identify and discard spurious features (such as background in images or graph size) that vary across domains but are not predictive.




\section{Methodology}\label{sec:methodology}

This section describes our key components and analysis framework for tackling OOD generalization on graphs. We first review the GPS layer, a hybrid architecture combining local message passing and global graph transformer attention for richer graph representations. Next, we discuss the strengths and limitations of GT and MPNN backbones for OOD tasks. We then review several \ac{DG} approaches aimed at improving robustness to unseen data distributions and employed in our experiments. Finally, to quantitatively assess OOD generalization in the latent feature space, we introduce two metrics that evaluate its quality. \ac{MMD}, which measures distributional shifts between ID and OOD embeddings, and the Silhouette score, which evaluates the clarity and separation of clusters formed by these representations. Together, they provide deeper and decoupled insights about the model's generalization abilities, given a specific dataset (or task).




\subsection{GPS layer: hybrid message-passing and graph-transformer}\label{sec:gps_layer}

The GPS layer is a hybrid \ac{MPNNs}+\ac{GT} architecture that enables both local and global reasoning on graphs by combining message passing with Transformer-style attention, as introduced in \cite{rampavsek2022recipe}. It updates node and edge features in parallel: a message passing neural network captures local neighborhood structure, computed as \( (X^{\ell+1}_M, E^{\ell+1}) = \text{MPNN}^{\ell}_e(X^{\ell}, E^{\ell}, A) \), while multi-head attention captures long-range dependencies by treating nodes as tokens and computing attention scores between all pairs using projected queries, keys, and values, computed as \( X^{\ell+1}_T = \text{MHA}^{\ell}(X^{\ell}) \). Both components are modular and can be instantiated with arbitrary local message-passing functions and global self-attention mechanisms, respectively. Their outputs are then fused through a multi-layer perceptron (MLP), which applies an element-wise sum followed by a learned transformation: \( X^{\ell+1} = \text{MLP}^{\ell}(X^{\ell}_M + X^{\ell}_T) \). The complete layer update is thus summarized as \( (X^{\ell+1}, E^{\ell+1}) = \text{GPS}^{\ell}(X^{\ell}, E^{\ell}, A) \). Edge features are processed only in the \ac{MPNNs} component. Residual connections, batch normalization \cite{ioffe2015batch}, dropout, and positional encoding are applied in practice but omitted from the equations for clarity (for more details refer to Section D.2 of L.~Rampášek et al.~\cite{rampavsek2022recipe}).

\subsection{Graph Transformer as a backbone for OOD tasks}

We hypothesize that a \ac{GT} possesses stronger generalization abilities than \ac{MPNNs}, potentially leading to improved OOD performance.

There are several reasons in favor of \ac{GT}.
First, \textit{global context modeling} is crucial for large graphs and tasks where the predicted label is casual to the entire or a large portion of the graph, as message-passing is computationally feasible only for two or three hops. 
Second, \textit{multi-head attention (MHA) with flexible receptive field}, which allows focusing on relevant graph parts, while ignoring irrelevant ones. Even though the MPNN variant, graph attention networks \cite{velivckovic2017graph}, may also be thought to possess those abilities, keep in mind that it only has a local receptive field. 
Third, \textit{MPNN's over-smoothing and over-squashing} can lead to information loss, compromise substructural integrity for larger diameter substructures, and insufficient edge information extraction. \ac{GT} mitigate these issues by aggregating information over all nodes in a graph, avoiding local structure bias and ensuring accurate node-edge relationships. 
Fourth, as shown by \cite{kreuzer2021rethinking}, \ac{GT} are more \textit{expressive} than \ac{MPNNs} bounded by 1-WL, because they can approximate any function using precomputed positional encodings based on the eigen decomposition of the graph Laplacian (universal approximation theorem \cite{cybenko1989approximation, hornik1991approximation}.)

\ac{GT}, while offering advantages in certain scenarios, might not always be the best choice for OOD tasks compared to \ac{MPNNs} for several reasons.
First, the \textit{over-globalizing problem} \cite{xing2024less} in \ac{GT} arises when nearby nodes hold most useful information. In such cases, local pattern modeling is crucial, while attention may overemphasize distant nodes (though \ac{GPS}, with its hybrid global-local modeling, may address this).
Second, the attention mechanism requires \textit{significant computational resources}, particularly when dealing with large graphs, though various “sparse” approaches have been proposed to mitigate this cost \cite{tay2022efficienttransformerssurvey}).
Third, \textit{virtual nodes} \cite{gilmer2017neural} for message-passing might be a simple alternative, facilitating global information exchange.
The above advantages and limitations of GT have been studied in the \ac{i.i.d} context, but lack an in-depth evaluation for \ac{OOD}.


\subsection{Domain generalization algorithms} 

Several methodological directions, which are detailed below, aim to tackle and improve the OOD challenge. 

\textit{Data augmentation approaches} generate more diverse training data, leading to better OOD generalization ability \cite{ding2022data}. 
These approaches include structure-wise augmentations, which potentially cover some unobserved testing topologies, feature-wise augmentations (by manipulating node features), and a combination of both. Specifically, Manifold-Mixup \cite{wang2021mixup} manipulates the training distribution in the semantic space by generating virtual graph representations.

\textit{Subgraph-based approaches} can offer more reliable graph predictions by encoding the local context of structures. One major area of research in causal-based graph models \cite{ chen2022learning} involves constructing a structural causal graph (SCM) for theoretical examination. Making sure the model is invariant to the non-causal element is a fundamental solution. 
\textit{Disentangled-based approaches} aim to learn representations that separate distinct and informative factors behind the graph data, process them, and characterize them by different mechanisms. \cite{fan2022debiasing}. 

\textit{Learning strategy approaches} focus on training schemes with specific objectives and constraints to promote \ac{OOD} generalization. Invariant learning aims to extract features or representations that remain consistent across domains during training \cite{wu2022discovering}. 
This often involves explicitly aligning feature distributions using techniques such as \ac{MMD}~\cite{li2018domain}, second-order correlation \cite{sun2016deep}, or methods like IRM and others \cite{arjovsky2019invariant, krueger2021out, sagawa2019distributionally}. In contrast, adversarial learning enforces domain invariance indirectly by training the feature extractor to fool a domain discriminator \cite{ganin2016domain}. 

In Section \ref{sec:experiments}, we show that in certain cases the backbone architecture choice is even more critical than \ac{DG} algorithms, as it determines feature extraction quality and significantly affects overall model performance.

\subsection{Squared Maximum Mean Discrepancy (MMD)}\label{sec:mmd}

To quantify alignment between domain distributions, the squared \ac{MMD} \cite{gretton2012kernel} distance is employed for the analysis of generalization abilities in Section~\ref{sec:rq3}. This process involves loading a trained model checkpoint, processing both ID and OOD test datasets to extract penultimate layer features, and then calculating the domain alignment using squared MMD.

The \ac{MMD} distance between two features sets $X = \{x_i\}_{i=1}^n$ and $Y = \{y_j\}_{j=1}^m$ is computed using a radial basis function (RBF) kernel based on the L1 distance:
\begin{equation}
k(x, y) = \exp\left(-\gamma \|x - y\|_1\right),
\end{equation}

The kernel bandwidth parameter $\gamma$ is set using the median-based method, i.e., as the inverse of the median pairwise L1 distance among all samples. The final squared \ac{MMD} distance (henceforth referred to simply as MMD) is defined by

\begin{equation}
\mathrm{MMD}^2(X, Y) = \frac{1}{n^2} \sum_{i,j=1}^n k(x_i, x_j)
+ \frac{1}{m^2} \sum_{i,j=1}^m k(y_i, y_j)
- \frac{2}{nm} \sum_{i=1}^n \sum_{j=1}^m k(x_i, y_j).
\end{equation}

\subsection{Silhouette score}\label{sec:silhouette}

The quality of clustering in a feature representation is quantified using the Silhouette score \cite{rousseeuw1987silhouettes} for the analysis of generalization abilities in Section~\ref{sec:rq3}. This process involves loading a trained model checkpoint, processing both ID and OOD test datasets to extract penultimate layer features, and then calculating the class-separation using the Silhouette score.

For a given sample $x_i$, the Silhouette score is defined as
\begin{equation}
s(i) = \frac{b(i) - a(i)}{\max\{a(i),\, b(i)\}},
\end{equation}
where $a(i)$ denotes the average distance from $x_i$ to all other points in the same cluster, and $b(i)$ is the minimum average distance from $x_i$ to all points in any other cluster. The overall Silhouette score is the mean of $s(i)$ over all samples.


\section{Experimental Results}\label{sec:experiments}

In this section, we systematically assess the OOD generalization abilities of \ac{GT} compared to \ac{MPNNs}. Our experiments are designed to provide a comprehensive evaluation by detailing the experimental setup, describing both accuracy-based and our proposed metrics results, and addressing the following research questions:
\begin{itemize}
    \item \textbf{RQ1.} How do vanilla GT and GPS architectures perform on OOD generalization tasks compared to MPNNs equipped with DG algorithms?

    \item \textbf{RQ2.} When equipped with DG algorithms, how do GT and GPS models compare to MPNNs in their OOD generalization performance?

    \item \textbf{RQ3.} How do different backbone architectures perform when evaluated using the proposed domain-alignment metric and class-separation metric? 

    \item \textbf{RQ4.} To what extent do our proposed metrics correlate with OOD accuracy and ID-OOD accuracy gap? Do they provide complementary insights into model generalization abilities?

    
    
    

\end{itemize}

\subsection{Experimental setup}

Experiments are conducted on six graph-classification datasets from the GOOD benchmark: CMNIST (color shift), Motif (size shift), Motif (basis shift), SST2 (length shift), HIV (size shift), and HIV (scaffold shift). 
For each dataset, covariate shift splits are applied based on domain features to create diverse distribution shifts between the training set, OOD validation set, and OOD test set. The training set is then shuffled and further divided into a final training set, ID validation set, and ID test set.
Figure \ref{fig:datasets} summarizes their key properties, such as input types, prediction tasks, and domain selection, and visually shows how training and test graph structures differ, illustrating the distribution shifts.

We evaluate a total of seven domain generalization (\ac{DG}) algorithms from the GOOD benchmark, including \ac{ERM} \cite{vapnik1999overview}, Invariant Risk Minimization (IRM) \cite{arjovsky2019invariant}, Variance Risk Extrapolation (VREx) \cite{krueger2021out}, Group Distributionally Robust Optimization (GroupDRO) \cite{sagawa2019distributionally}, Domain-Adversarial Neural Networks (DANN) \cite{ganin2016domain}, and Correlation Alignment (Coral) \cite{sun2016deep}, and Mixup \cite{wang2021mixup}. 
 Each DG algorithm has been modified to support each backbone. 

For the backbone architectures, we use \ac{vGIN}, \ac{MHA}, and GPS. Our baseline is vGIN, an MPNN variant that extends GIN with a virtual node to enhance global information flow, consistent with the backbone choice in the GOOD benchmark (except for Motif tasks, where we used GIN instead of vGIN). For the pure GT backbone, MHA, which applies a global attention mechanism between graph nodes, is used. We selected \ac{RW} embeddings \cite{dwivedi2021graph} as positional encodings, based on their demonstrated effectiveness on ID tasks \cite{tang2025opengt}. For the hybrid MPNN-GT backbone, we use GPS, as described in Section \ref{sec:gps_layer}. Note that GPS uses standard GIN, rather than vGIN, for local message passing and MHA-RW for the global receptive field. All models are trained using the hyperparameters listed in Table~\ref{tab:hyperparams}, with the number of trainable parameters matched across backbones (per-task) as closely as possible to ensure a fair comparison. Global mean pooling is applied for node-level aggregation. Additional details are provided in Appendix \ref{sec:exp_details}.

\subsection{Evaluation metrics}

While OOD accuracy, measured on the OOD test set, is the most prevalent metric for quantifying OOD performance, it alone does not capture all aspects of generalization such as accuracy degradation between the \ac{ID} to \ac{OOD} settings. To address this, we also report the ID-OOD accuracy gap, defined as $gap = ID_{test} \text{-}    OOD_{test}$, which is a common metric that is used in \ac{NLP} and computer-vision \cite{hendrycks2020pretrained, zhang2022delving}. Referred to simply as the accuracy gap throughout the paper, this metric reflects the model’s sensitivity to distribution shifts relative to its ID performance and complements the OOD accuracy metric.

To assess the model's generalization capabilities beyond standard accuracy metrics, we evaluated the squared \ac{MMD} distance between the penultimate-layer representations of \ac{ID} and \ac{OOD} test datasets. This provides a quantitative measure of feature space alignment across domains. Furthermore, we computed the Silhouette score on the concatenated penultimate-layer features from both \ac{ID} and \ac{OOD} datasets to quantify how well each data point fits within its assigned cluster (intra-cluster distance) compared to other clusters (inter-cluster distance).

The best checkpoint is selected based on the \ac{ID} validation set, reflecting realistic scenarios where \ac{OOD} data is unavailable during training or validation. This contrasts with prior works that select checkpoints using an \ac{OOD} validation set, inflating performance beyond realistic deployment conditions. Final evaluation is then conducted on both \ac{ID} and \ac{OOD} test sets.


\subsection{Vanilla GT/GPS versus MPNNs equipped with DG algorithms (RQ1)}

Figure~\ref{fig:ood-bar} clearly shows that GPS (green) achieves the highest average \ac{OOD} accuracy across benchmarks, even when trained only with standard \ac{ERM} (without explicit \ac{DG} algorithms). In contrast, \ac{vGIN} (blue), although equipped with \ac{DG} algorithms, underperforms compared to both GPS and MHA on CMNIST, Motif, and SST tasks. While vGIN leads on both HIV's tasks, it can't be seen in Figure~\ref{fig:ood-bar} but can be seen in Table~\ref{tab:ood_results}, this gap narrows down in the HIV-size but not in HIV-scaffold. vGIN leads on both HIV tasks, however, since Figure~\ref{fig:ood-bar} reflects vGIN equipped with DG algorithms while GPS and MHA are vanilla, the actual gap narrows on HIV-size but remains pronounced on HIV-scaffold, as shown in Table~\ref{tab:ood_results}. Particularly on the challenging CMNIST-color task, GPS stands out, whereas \ac{MHA} (red) slightly surpasses it on Motif tasks but does not match its overall robustness across all benchmarks. For example, GPS outperforms \ac{vGIN} by a large margin on CMNIST-color (87\% vs. 29.8\%) and remains highly competitive on all other benchmarks, with an average \ac{OOD} accuracy of 67.3\% compared to 52\% for \ac{vGIN} and 56.7\% for \ac{MHA}.

These results underscore the benefits of using \ac{GT} for improving \ac{OOD} robustness (as also observed in computer vision \cite{zhang2022delving,sultana2022self}). The ability of global attention and context modeling to capture long-range dependencies appears to significantly boost OOD accuracy under distribution shifts on those tasks. Notably, hybrid architectures such as GPS, which integrate local message passing with global attention, further amplify this advantage. At the same time, results on specific tasks such as HIV-scaffold indicate that in certain shifts, local message passing alone may be better suited. These findings indicate that incorporating \ac{GT}, particularly in a hybrid backbone, represents a promising direction for advancing \ac{OOD} generalization in graph learning.

\subsection{GT/GPS versus MPNNs when equipped with DG algorithms (RQ2)}

Table~\ref{tab:ood_results} benchmarks three backbone architectures (vGIN, MHA, and GPS) across six distribution-shifted tasks from the GOOD benchmark under seven \ac{DG} algorithms. While vGIN often performs well in-domain, it struggles under distribution shifts, at four out of six tasks.

In the CMNIST with color shift task, IRM combined with GPS achieves the best OOD accuracy (89.26\%) and the near-lowest accuracy gap (0.68). GPS demonstrates strong invariance to feature-based distribution shifts, in contrast to vGIN and MHA, which struggle to generalize when used alone.

For the Motif with size shift task, MHA achieves the highest OOD accuracy when combined with GroupDRO, while GPS lags by approximately 5\%. This suggests that size-based generalization may favor pure Transformers, likely due to their inherently long-range receptive fields.

In the Motif with basis shift task, MHA and GPS exhibit closely matched OOD accuracy and gap, suggesting that both attention-based and hybrid backbones are well-suited to handling structural shifts in graph topology.

For the SST2 with length shift task, GPS achieves the highest OOD accuracy (83.29\%) with GroupDRO, while also maintaining low accuracy gaps across DG algorithms. Although vGIN attains relatively strong OOD performance here, the gains from GPS remain consistent and underline its robustness.

Last for the HIV tasks, when comparing the strongest OOD variants of each backbone, vGIN leads by roughly 2\% on the size shift and 5\% on the scaffold shift, suggesting that locality in message passing can be better suited for certain distribution shifts.

\begin{figure}[H]
    \centering
    \begin{subfigure}[t]{0.44\linewidth}
        \centering
        \includegraphics[width=\linewidth]{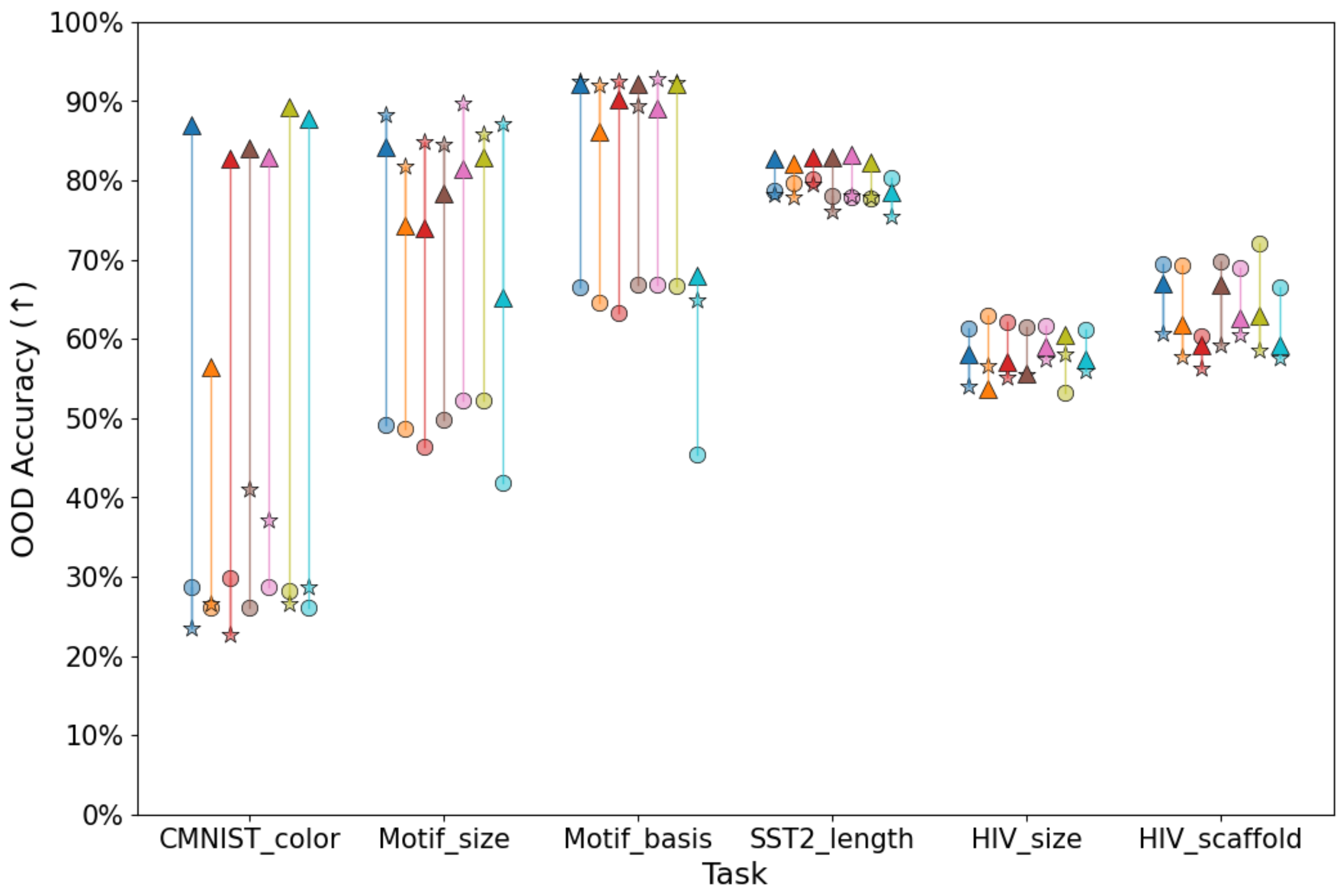}
        \caption{OOD test accuracy}
        \label{fig:ood_acc}
    \end{subfigure}
    \hfill
    \begin{subfigure}[t]{0.55\linewidth}
        \centering
        \includegraphics[width=\linewidth]{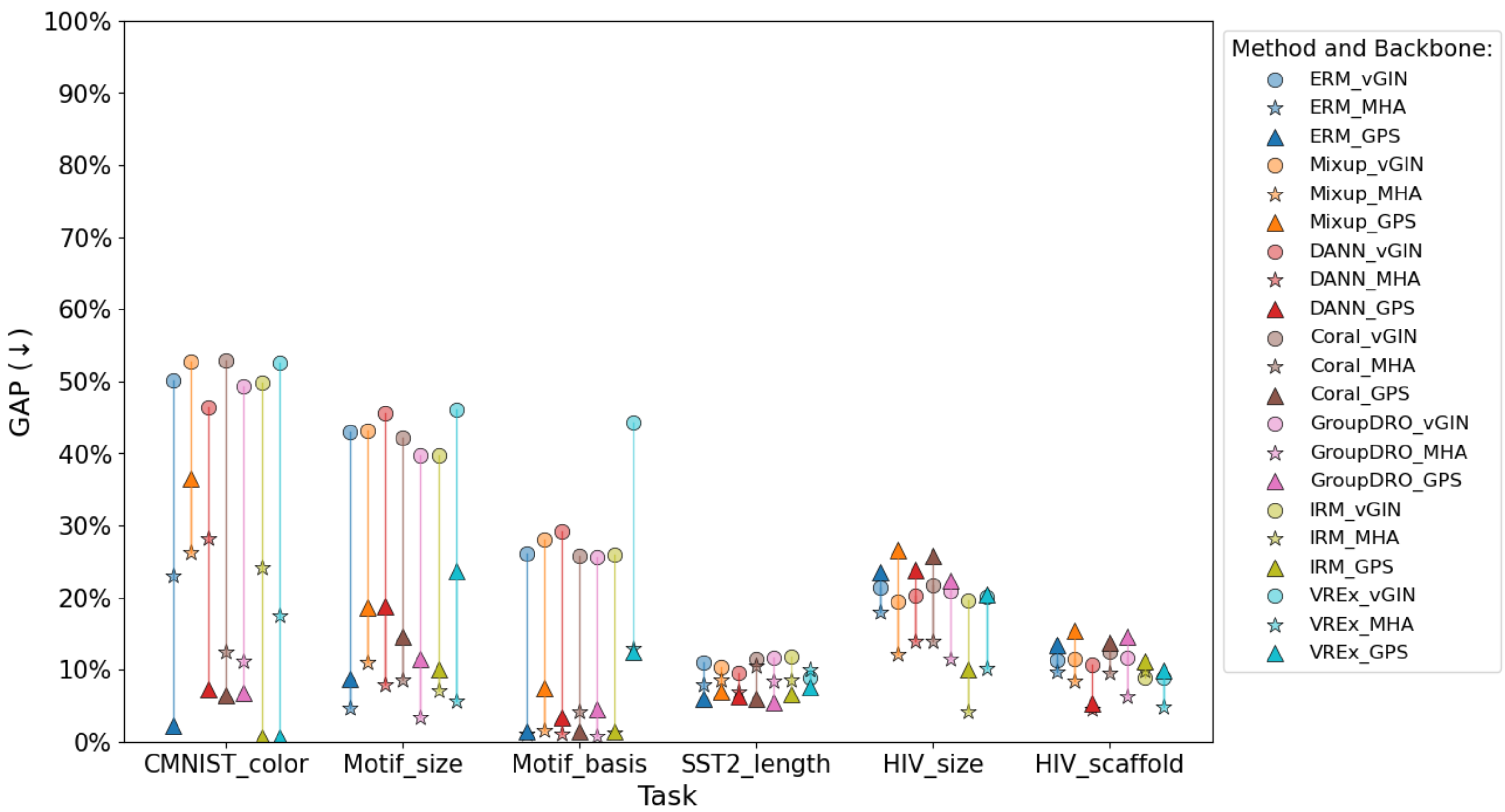}
        \caption{ID-OOD accuracy gap}
        \label{fig:gap}
    \end{subfigure}
    \caption{Visualization of accuracy results reported in Table \ref{tab:ood_results}. Each column corresponds to a different benchmark, and each color indicates a distinct DG algorithm. Marker shapes represent the backbone: circles for vGIN, stars for MHA, and triangles for GPS.}
    \label{fig:ood_acc_and_gap}
\end{figure}

Across four out of six graph classification tasks and various DG algorithms, GPS consistently achieves the highest or near-highest \ac{OOD} accuracies and smallest accuracy GAP, demonstrating the advantage of combining local message passing with global attention. In contrast, vGIN exhibits large performance drops (except for the HIV tasks), especially on CMNIST and Motif—where the accuracy gaps exceed 25 to 50 points, highlighting its sensitivity to distribution shifts. Meanwhile, MHA perform well on Motif tasks but less effectively on others. These patterns suggest that GPS offers a more reliable backbone for maintaining performance under most distribution shifts. The results are visualized in Figure~\ref{fig:ood_acc_and_gap}, where the left panel shows the OOD test accuracy results, and the right one presents the gap accuracies. ID results are discussed in Appendix \ref{sec:additional_exp}.

\begin{table}[htbp]
\centering
\tiny
\setlength{\tabcolsep}{0.4pt}
\begin{tabularx}{\textwidth}{ll|ccc|ccc|ccc|ccc|ccc|ccc|}
\toprule
\textbf{DG Alg.} & \textbf{Backbone}
& \multicolumn{3}{c|}{\textbf{CMNIST-color}}
& \multicolumn{3}{c|}{\textbf{Motif-size}}
& \multicolumn{3}{c|}{\textbf{Motif-basis}}
& \multicolumn{3}{c|}{\textbf{SST2-length}}
& \multicolumn{3}{c|}{\textbf{HIV-size}}
& \multicolumn{3}{c|}{\textbf{HIV-scaffold}} \\
\cmidrule(lr){3-5} \cmidrule(lr){6-8} \cmidrule(lr){9-11} \cmidrule(lr){12-14} \cmidrule(lr){15-17} \cmidrule(lr){18-20}
& & ID acc. & OOD acc. & Gap 
  & ID acc. & OOD acc. & Gap 
  & ID acc. & OOD acc. & Gap 
  & ID acc. & OOD acc. & Gap 
  & ID acc. & OOD acc. & Gap 
  & ID acc. & OOD acc. & Gap \\
\midrule
    ERM & vGIN  & 78.84 & 28.74 & 50.11 & 92.15 & 49.21 & 42.94 & 92.63 & 66.59 & 26.05 & 89.65 & 78.61 & 11.04 & 82.65 & 61.34 & 21.30 & 80.77 & 69.51 & 11.26 \\
    ERM & MHA   & 46.38 & 23.43 & 22.95 & 92.97 & 88.34 & 4.63  & \textbf{93.62} & 92.48 & 1.13  & 86.20 & 78.24 & 7.97  & 71.86 & 53.94 & 17.92 & 70.24 & 60.61 & 9.63  \\
    ERM & GPS   & 89.32 & 87.03 & 2.29  & 92.96 & 84.24 & 8.72  & 93.54 & 92.17 & 1.38  & 88.71 & 82.80 & 5.92  & 81.55 & 58.05 & 23.51 & 80.56 & 67.05 & 13.51 \\
\midrule
    Mixup & vGIN & 78.80 & 26.05 & 52.75 & 91.80 & 48.60 & 43.20 & 92.62 & 64.56 & 28.06 & \textbf{90.02} & 79.66 & 10.36 & 82.37 & \textbf{62.95} & 19.42 & 80.72 & 69.22 & 11.50 \\
    Mixup & MHA  & 52.88 & 26.58 & 26.30 & 92.74 & 81.73 & 11.01 & 93.58 & 92.03 & 1.56  & 86.47 & 77.86 & 8.61  & 68.83 & 56.67 & 12.16 & 66.08 & 57.69 & 8.39  \\
    Mixup & GPS  & \textbf{92.81} & 56.41 & 36.41 & 92.88 & 74.31 & 18.58 & 93.57 & 86.17 & 7.40  & 89.06 & 82.14 & 6.91  & 80.30 & 53.67 & 26.63 & 77.10 & 61.77 & 15.33 \\
\midrule
    DANN & vGIN  & 76.24 & 29.85 & 46.39 & 91.99 & 46.39 & 45.60 & 92.58 & 63.33 & 29.25 & 89.76 & 80.20 & 9.56  & 82.34 & 62.08 & 20.25 & 70.93 & 60.34 & 10.60 \\
    DANN & MHA   & 50.91 & 22.68 & 28.23 & 92.90 & 84.94 & 7.96  & 93.59 & 92.56 & 1.03  & 86.47 & 79.52 & 6.96  & 69.08 & 55.21 & 13.87 & 60.80 & 56.36 & \textbf{4.45}  \\
    DANN & GPS   & 89.95 & 82.69 & 7.26  & 92.81 & 74.02 & 18.79 & 93.60 & 90.29 & 3.31  & 89.14 & 82.87 & 6.27  & 80.99 & 57.15 & 23.84 & 64.54 & 59.28 & 5.26  \\
\midrule
    Coral & vGIN & 78.94 & 26.09 & 52.85 & 92.07 & 49.83 & 42.24 & 92.64 & 66.80 & 25.84 & 89.49 & 78.05 & 11.44 & \textbf{83.08} & 61.41 & 21.66 & \textbf{82.12} & 69.72 & 12.40 \\
    Coral & MHA  & 53.57 & 41.10 & 12.47 & \textbf{92.99} & 84.46 & 8.53  & 93.60 & 89.41 & 4.19  & 86.66 & 76.15 & 10.51 & 69.38 & 55.43 & 13.95 & 68.81 & 59.25 & 9.56  \\
    Coral & GPS  & 90.53 & 84.03 & 6.49  & 92.84 & 78.33 & 14.52 & 93.52 & 92.18 & 1.34  & 88.90 & 82.93 & 5.97  & 81.35 & 55.64 & 25.71 & 80.47 & 66.79 & 13.68 \\
\midrule
    GroupDRO & vGIN & 78.06 & 28.71 & 49.35 & 91.99 & 52.24 & 39.76 & 92.60 & 66.92 & 25.68 & 89.53 & 77.83 & 11.70 & 82.64 & 61.67 & 20.97 & 80.61 & 68.92 & 11.69 \\
    GroupDRO & MHA  & 48.25 & 37.10 & 11.15 & 92.96 & \textbf{89.67} & \textbf{3.29} & 93.58 & \textbf{92.78} & \textbf{0.80} & 86.33 & 77.99 & 8.35  & 68.96 & 57.44 & 11.53 & 66.84 & 60.52 & 6.32  \\
    GroupDRO & GPS  & 89.70 & 82.96 & 6.74  & 92.87 & 81.39 & 11.48 & 93.61 & 89.14 & 4.47  & 88.72 & \textbf{83.29} & \textbf{5.43}  & 81.37 & 59.04 & 22.33 & 77.06 & 62.56 & 14.50 \\
\midrule
    IRM & vGIN   & 77.99 & 28.13 & 49.86 & 91.97 & 52.20 & 39.77 & 92.62 & 66.62 & 26.01 & 89.47 & 77.68 & 11.80 & 72.81 & 53.23 & 19.58 & 80.90 & \textbf{72.06} & 8.84  \\
    IRM & MHA   & 50.74 & 26.61 & 24.13 & \textbf{92.99} & 85.82 & 7.18  & 93.59 & 92.34 & 1.25  & 86.48 & 77.89 & 8.59  & 62.26 & 58.14 & \textbf{4.12}  & 68.38 & 58.57 & 9.81  \\
    IRM & GPS   & 89.95 & \textbf{89.26} & 0.68 & 92.93 & 82.93 & 10.00 & 93.60 & 92.24 & 1.36  & 88.89 & 82.23 & 6.66  & 70.59 & 60.58 & 10.02 & 74.01 & 62.89 & 11.12 \\
\midrule
    VREx & vGIN  & 78.54 & 26.04 & 52.50 & 87.82 & 41.79 & 46.03 & 89.65 & 45.38 & 44.28 & 89.17 & 80.32 & 8.85  & 81.31 & 61.15 & 20.16 & 75.47 & 66.51 & 8.96  \\
    VREx & MHA   & 46.28 & 28.76 & 17.52 & 92.71 & 87.09 & 5.63  & 77.87 & 64.95 & 12.92 & 85.43 & 75.48 & 9.95  & 66.20 & 56.03 & 10.17 & 62.50 & 57.66 & 4.84  \\
    VREx & GPS   & 88.28 & 87.72 & \textbf{0.56} & 88.87 & 65.14 & 23.73 & 80.47 & 67.97 & 12.50 & 86.25 & 78.58 & 7.67  & 77.92 & 57.46 & 20.46 & 69.10 & 59.23 & 9.87  \\
\bottomrule
\end{tabularx}
\vspace{1pt}
\caption{Comparison of \ac{vGIN}, \ac{MHA}, and GPS backbones' ID/OOD accuracies and accuracy gap across GOOD's graph-level classification tasks. Results are reported for different backbone architectures and DG algorithms. Bold values highlight the best-performing method in each column.}
\label{tab:ood_results}
\end{table}

\subsection{Quantification of model generalization abilities (RQ3)}\label{sec:rq3}

\begin{figure}[t]
    \centering
    \begin{subfigure}[t]{0.435\linewidth}
        \centering
        \includegraphics[width=\linewidth]{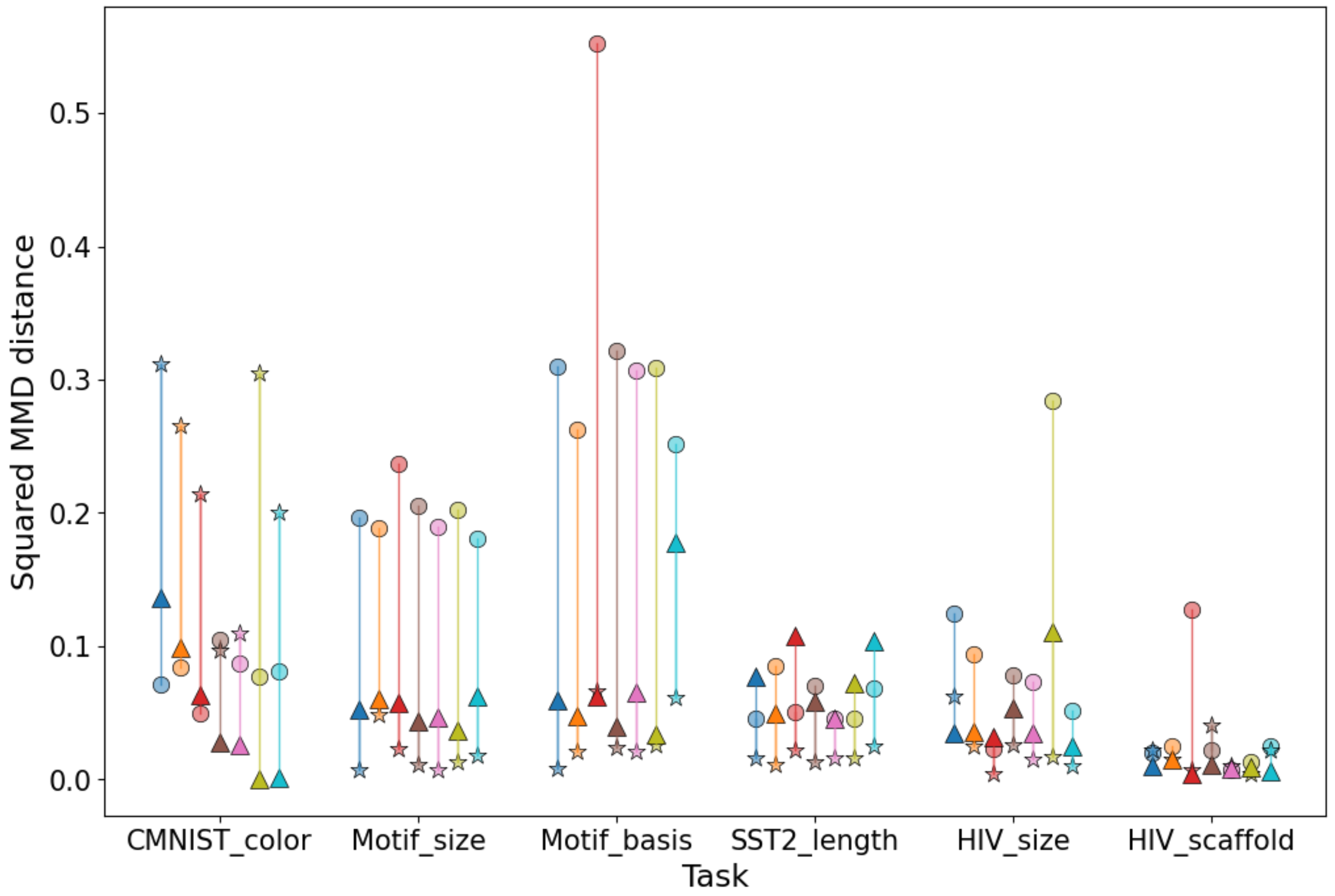}
        \caption{\ac{MMD} distance ($\downarrow$).}
        \label{fig:mmd}
    \end{subfigure}
    \hfill
    \begin{subfigure}[t]{0.555\linewidth}
        \centering
        \includegraphics[width=\linewidth]{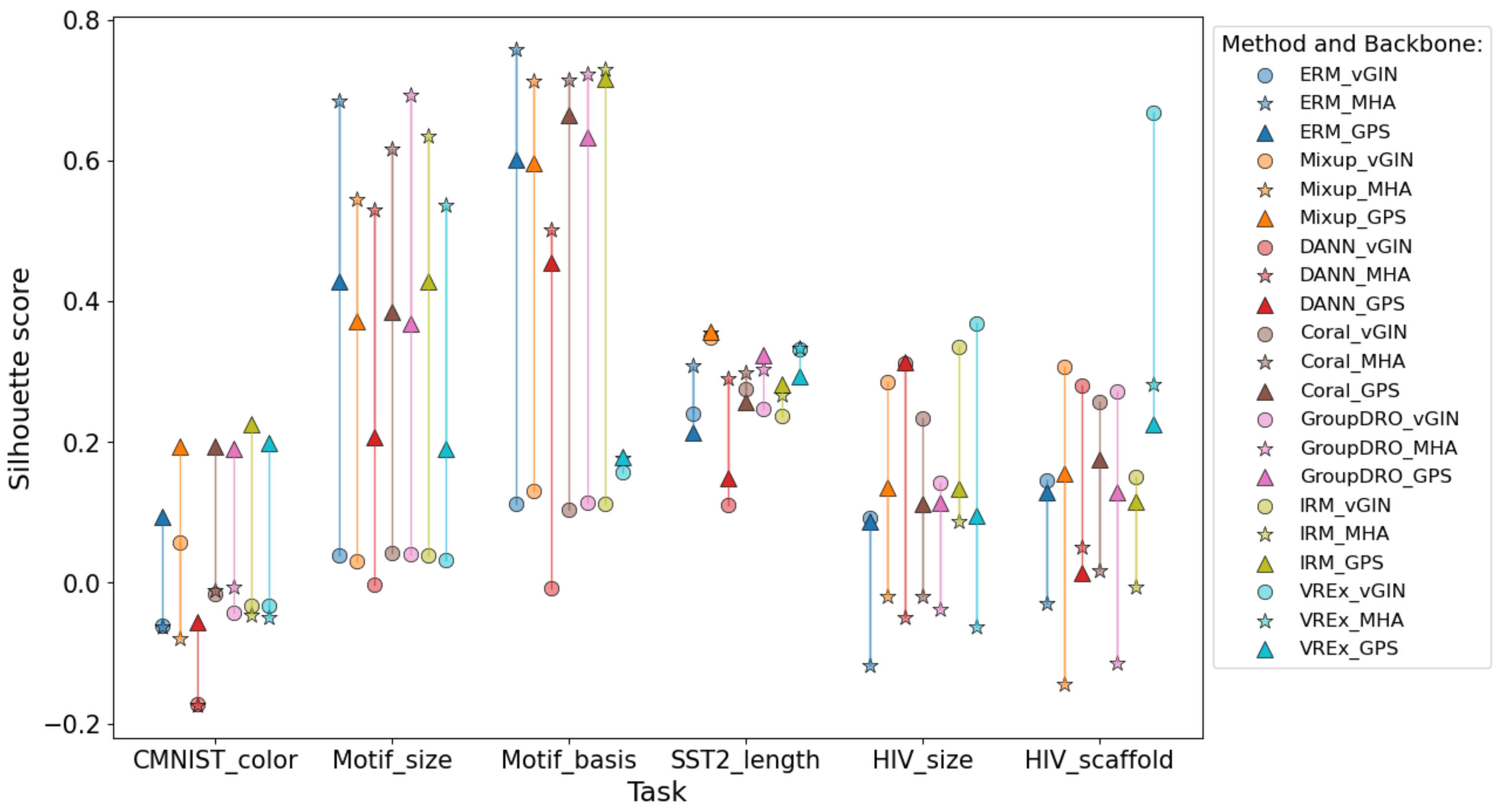}
        \caption{Silhouette score ($\uparrow$).}
        \label{fig:silhouette}
    \end{subfigure}
    \caption{Visualization of the proposed metrics results in Table \ref{tab:mmd_silhouette_results} is based on penultimate-layer features extracted from ID and OOD datasets. Each column corresponds to a different benchmark, and each color indicates a distinct DG algorithm. Marker shapes represent the backbone: circles for vGIN, stars for MHA, and triangles for GPS.}
    \label{fig:mmd_and_silhouette}
\end{figure}

Figure~\ref{fig:mmd_and_silhouette} presents a quantitative evaluation of generalization abilities in the semantic space across \ac{DG} algorithms and model backbones using the two proposed metrics: \ac{MMD} distance (domain-alignment metric) and Silhouette Score (class-separation metric). 

The MMD metric (Figure~\ref{fig:mmd}) quantifies the distributional similarity between penultimate-layer features of ID and OOD test sets; values closer to zero indicate greater invariance to domain shift in the learned latent representations. The Silhouette Score in Figure~\ref{fig:silhouette} is computed over the union of ID and OOD features to assess whether the model’s representations maintain coherent class-wise clustering under domain shift. A closer to one Silhouette score indicates both high intra-cluster and low inter-cluster similarities, suggesting strong semantic generalization.  This complements MMD distance, which measures domain invariance but may not be sensitive enough to the semantic structure. Together, these metrics reveal whether the model’s generalization arises from features that are domain-invariant and class-consistent, allowing class structure to remain separable even under domain shift.

Table~\ref{tab:mmd_silhouette_results} shows notable differences across backbones and DG algorithms, with backbone choice often exerting a stronger influence than the algorithm itself. Interestingly, ERM paired with MHA or GPS frequently matches or outperforms specialized DG approaches. At the same time, performance is highly dataset-dependent. On CMNIST-color, GPS achieves the best Silhouette score (0.216 with IRM) and lowest MMD distance (0.001, tied across multiple algorithms), while MHA and vGIN produce negative Silhouette scores for all DG methods, indicating poor domain separation. For both Motif tasks, MHA ranks highest (Silhouette score $>0.68$ even with ERM; MMD distance $<0.08$), while GPS also performs well across algorithms, and vGIN struggles consistently. For SST2-length, GPS with Mixup achieves the best Silhouette score (0.356), and MHA with Mixup reaches the lowest MMD distance (0.012). For both HIV tasks, vGIN have the highest Silhouette scores, but MHA have the lowest MMD distances. These trends (on four out of six benchmarks) underscore that the architectural design of MHA and GPS leads to more robust, tighter clusters and domain-aligned representations, often surpassing what DG-specific methods can achieve with weaker backbones. Additionally, we observe that optimizing for Silhouette score and MMD distance does not always coincide, highlighting the importance of evaluating these two complementary aspects, clustering and alignment, when assessing domain generalization.

\begin{figure}[t]
    \centering
    \begin{subfigure}[t]{0.31\linewidth}
        \centering
        \includegraphics[width=\linewidth]{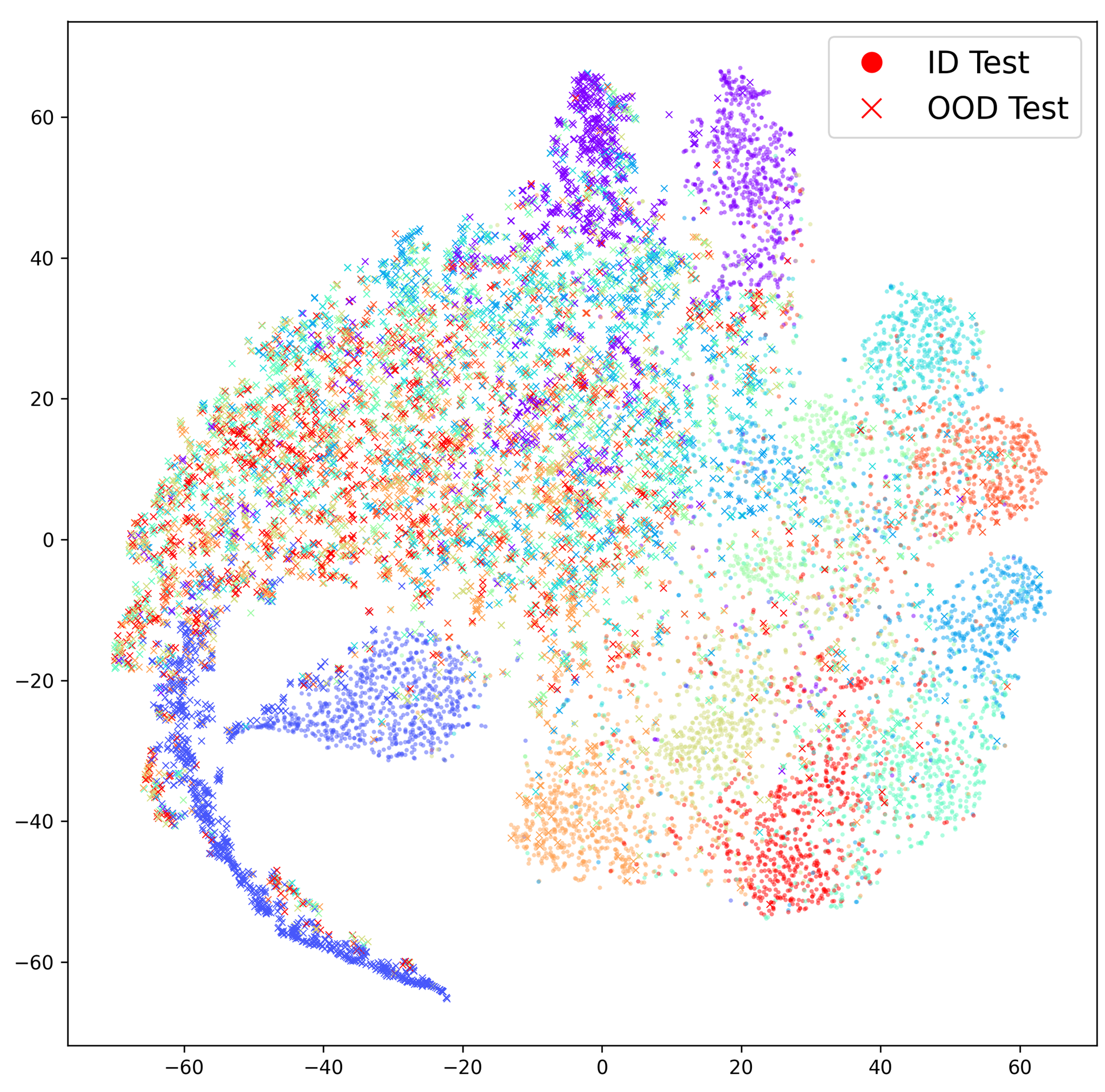}
        \caption{\ac{vGIN}}
        \label{fig:vgin_vis}
    \end{subfigure}
    \hfill
    \begin{subfigure}[t]{0.31\linewidth}
        \centering
        \includegraphics[width=\linewidth]{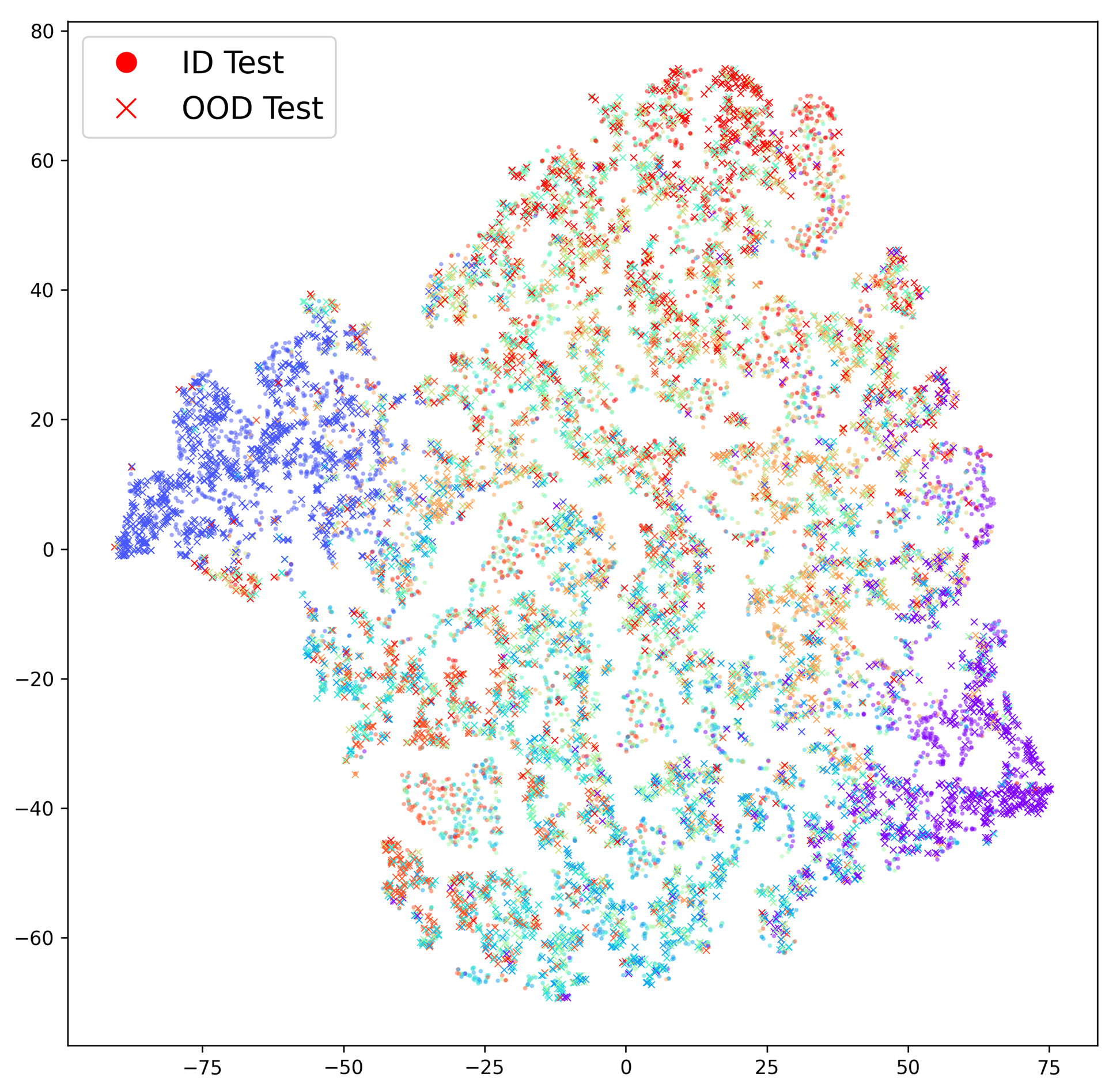}
        \caption{\ac{MHA}}
        \label{fig:mha_vis}
    \end{subfigure}
    \hfill
    \begin{subfigure}[t]{0.36\linewidth}
        \centering
        \includegraphics[width=\linewidth]{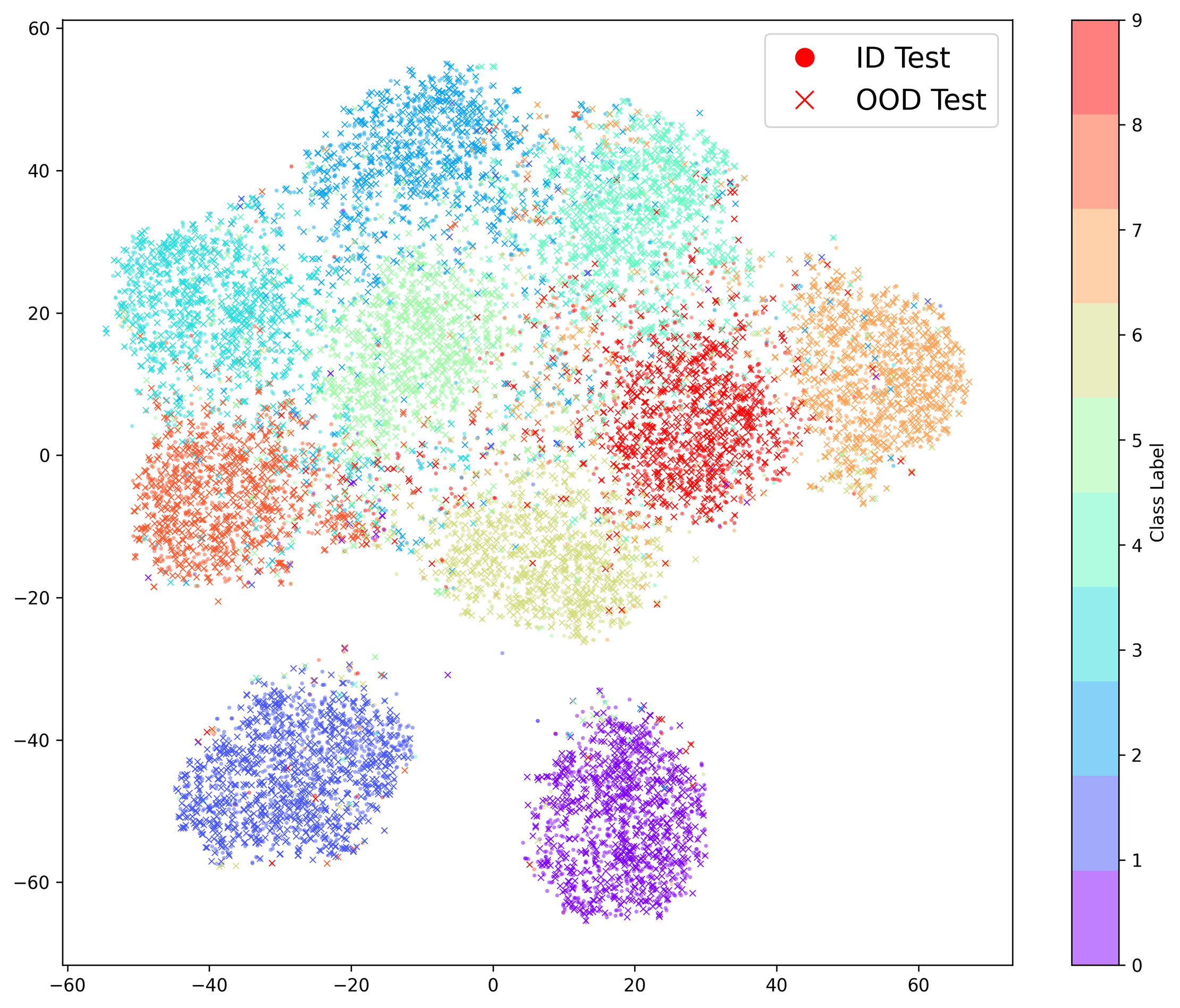}
        \caption{GPS}
        \label{fig:gps_vis}
    \end{subfigure}
    \caption{\ac{t-SNE} visualization of penultimate-layer feature of three backbones (vGIN, MHA, and GPS) with ERM on CMNIST-color. Colors correspond to the 10 class labels. Circles represent ID test samples; crosses represent OOD test samples.}
    \label{fig:t-sne}
\end{figure}

Figure~\ref{fig:t-sne} provides a qualitative visualization of the learned penultimate-layer features for the CMNIST-color dataset, using \ac{t-SNE} to project embeddings from each backbone (with ERM) into a 2D space. Each point represents a test sample, with circles denoting ID data and crosses indicating OOD data. The GPS backbone (Figure~\ref{fig:gps_vis}) produces well-separated, compact class clusters where ID and OOD samples generally overlap within the same regions, suggesting robust class semantics and domain-invariant features. MHA (Figure~\ref{fig:mha_vis}) shows moderate clustering, but with greater class dispersion and visible shifts between ID and OOD points. Finally, vGIN (Figure~\ref{fig:vgin_vis}) yields entangled and diffuse representations with poor class separation and pronounced domain shift. These qualitative patterns align closely with the Silhouette scores and MMD distances reported in Table~\ref{tab:mmd_silhouette_results} for CMNIST-color, where GPS outperforms both MHA and vGIN, highlighting its ability to learn both discriminative and domain-consistent representations even without explicit domain generalization techniques.

\begin{table}[t]
\centering
\tiny
\begin{tabular}{ll|cc|cc|cc|cc|cc|cc}
\toprule
\textbf{DG Alg.} & \textbf{Backbone}
& \multicolumn{2}{c|}{\textbf{CMNIST-color}}
& \multicolumn{2}{c|}{\textbf{Motif-size}}
& \multicolumn{2}{c|}{\textbf{Motif-basis}}
& \multicolumn{2}{c|}{\textbf{SST2-length}}
& \multicolumn{2}{c|}{\textbf{HIV-size}}
& \multicolumn{2}{c}{\textbf{HIV-scaffold}} \\
\cmidrule(lr){3-4} \cmidrule(lr){5-6} \cmidrule(lr){7-8} \cmidrule(lr){9-10} \cmidrule(lr){11-12} \cmidrule(lr){13-14}
& & Silh. & MMD & Silh. & MMD & Silh. & MMD & Silh. & MMD & Silh. & MMD & Silh. & MMD \\
\midrule
\multirow{3}{*}{ERM}
& vGIN & -0.060 & 0.071 & 0.038 & 0.197 & 0.111 & 0.310 & 0.240 & 0.046 & 0.092 & 0.124 & 0.145 & 0.020 \\
& MHA  & -0.062 & 0.312 & 0.685 & \textbf{0.007} & \textbf{0.758} & \textbf{0.008} & 0.309 & 0.016 & -0.117 & 0.062 & -0.029 & 0.022 \\
& GPS  & 0.094 & 0.136 & 0.429 & 0.053 & 0.602 & 0.060 & 0.213 & 0.077 & 0.087 & 0.035 & 0.128 & 0.010 \\
\midrule
\multirow{3}{*}{Mixup}
& vGIN & 0.058 & 0.084 & 0.031 & 0.189 & 0.130 & 0.263 & 0.348 & 0.086 & 0.284 & 0.094 & 0.307 & 0.025 \\
& MHA  & -0.080 & 0.266 & 0.545 & 0.049 & 0.712 & 0.021 & 0.354 & \textbf{0.012} & -0.020 & 0.025 & -0.145 & 0.015 \\
& GPS  & 0.194 & 0.099 & 0.371 & 0.061 & 0.596 & 0.048 & \textbf{0.356} & 0.050 & 0.135 & 0.036 & 0.156 & 0.015 \\
\midrule
\multirow{3}{*}{DANN}
& vGIN & -0.173 & 0.050 & -0.003 & 0.236 & -0.008 & 0.553 & 0.111 & 0.051 & 0.312 & 0.023 & 0.281 & 0.127 \\
& MHA  & -0.174 & 0.214 & 0.530 & 0.023 & 0.502 & 0.066 & 0.290 & 0.022 & -0.050 & \textbf{0.005} & 0.051 & 0.008 \\
& GPS  & -0.056 & 0.064 & 0.207 & 0.057 & 0.455 & 0.062 & 0.148 & 0.108 & 0.313 & 0.032 & 0.013 & 0.005 \\
\midrule
\multirow{3}{*}{Coral}
& vGIN & -0.017 & 0.105 & 0.041 & 0.205 & 0.104 & 0.322 & 0.276 & 0.070 & 0.233 & 0.078 & 0.258 & 0.022 \\
& MHA  & -0.011 & 0.097 & 0.616 & 0.011 & 0.714 & 0.024 & 0.299 & 0.013 & -0.019 & 0.026 & 0.018 & 0.041 \\
& GPS  & 0.193 & 0.028 & 0.386 & 0.044 & 0.665 & 0.040 & 0.257 & 0.059 & 0.113 & 0.054 & 0.176 & 0.011 \\
\midrule
\multirow{3}{*}{GroupDRO}
& vGIN & -0.043 & 0.088 & 0.041 & 0.189 & 0.113 & 0.307 & 0.246 & 0.046 & 0.142 & 0.074 & 0.272 & 0.007 \\
& MHA  & -0.006 & 0.110 & \textbf{0.693} & \textbf{0.007} & 0.723 & 0.021 & 0.304 & 0.016 & -0.038 & 0.015 & -0.115 & 0.010 \\
& GPS  & 0.190 & 0.026 & 0.368 & 0.047 & 0.633 & 0.066 & 0.324 & 0.046 & 0.115 & 0.035 & 0.128 & 0.009 \\
\midrule
\multirow{3}{*}{IRM}
& vGIN & -0.032 & 0.077 & 0.039 & 0.202 & 0.112 & 0.309 & 0.237 & 0.046 & 0.336 & 0.284 & 0.150 & 0.013 \\
& MHA  & -0.045 & 0.305 & 0.635 & 0.013 & 0.730 & 0.026 & 0.267 & 0.016 & 0.088 & 0.017 & -0.006 & \textbf{0.004} \\
& GPS  & \textbf{0.226} & \textbf{0.001} & 0.428 & 0.037 & 0.716 & 0.034 & 0.282 & 0.072 & 0.133 & 0.110 & 0.116 & 0.009 \\
\midrule
\multirow{3}{*}{VREx}
& vGIN & -0.033 & 0.081 & 0.032 & 0.181 & 0.156 & 0.252 & 0.332 & 0.068 & \textbf{0.368} & 0.051 & \textbf{0.667} & 0.025 \\
& MHA  & -0.050 & 0.200 & 0.536 & 0.018 & 0.177 & 0.061 & 0.334 & 0.025 & -0.063 & 0.011 & 0.282 & 0.022 \\
& GPS  & 0.198 & \textbf{0.001} & 0.190 & 0.062 & 0.179 & 0.178 & 0.293 & 0.104 & 0.096 & 0.025 & 0.225 & 0.006 \\
\bottomrule
\end{tabular}
\vspace{1pt}
\caption{Comparison of models using the proposed Silhouette score and squared MMD distance across the graph-level classification tasks in the GOOD benchmark. Results are reported for different backbone architectures and DG algorithms. Bold values highlight the best-performing method in each column.}
\label{tab:mmd_silhouette_results}
\end{table}

\subsection{Complementary abilities of model evaluation and generalization metrics (RQ4)}\label{sec:rq4}

Although OOD accuracy, ID-OOD accuracy gap, MMD distance, and Silhouette score often trend in the same direction, they probe different aspects of generalization and therefore do not always correlate tightly. Models that attain similar OOD accuracy can differ markedly in the way they organize their latent representations. 
When two models achieve the same high OOD accuracy, a larger gap indicates higher ID performance, suggesting greater potential for further improving generalization, though it may also carry some risk if future OOD shifts differ from the current test set, highlighting why both OOD accuracy and the gap metric are needed to assess real-world deployment suitability.

Moreover, selecting a model solely based on OOD accuracy and accuracy gap can be misleading. For instance, in the Motif-basis task, MHA-ERM exhibits OOD accuracy and gap values similar to other variants; however, models such as MHA-DANN yield a substantially lower Silhouette score, indicating poorly separated class clusters, and a higher MMD distance, reflecting misaligned domain feature distributions, underscoring the need for complementary metrics beyond simple accuracy measures.

In most cases, as intuition suggests, high OOD accuracy is typically accompanied by a small accuracy gap. It is also associated with a low MMD distance, which indicates well-aligned domains, and a high Silhouette score, which reflects well-separated class clusters. Yet the counter examples above illustrate that these regularities are not universal. Reporting all six metrics offers a more comprehensive and interpretable understanding of each model’s generalization capabilities. As a result, better-informed decisions can be made when selecting models for real-world applications.

\section{Conclusion}\label{sec:conclusion}

This work presents the first systematic evaluation of Graph Transformer (GT) backbones for out-of-distribution (OOD) generalization in supervised graph classification. Our results demonstrate that GraphGPS (GPS), a hybrid architecture integrating local message passing with global transformer attention, consistently outperforms both MPNN (vGIN) and pure attention-based (MHA) models (on four out of six benchmarks), demonstrating the importance of backbone design for OOD robustness. Notably, GPS achieves an average OOD accuracy of 67.3\% across benchmarks—substantially higher than MHA (56.7\%) and vGIN (52\%), even when only the latter is equipped with the strongest domain generalization algorithms. On challenging shifts such as CMNIST-color, GPS attains 87\% OOD accuracy compared to vGIN’s 29.8\%.

Crucially, we show that relying solely on accuracy-based metrics can be misleading when assessing generalization abilities. Our proposed post-hoc analysis, leveraging Maximum Mean Discrepancy (MMD) for domain-alignment and Silhouette score for class-separation, reveals that models with similar OOD accuracy may differ significantly in latent representation quality. For instance, some models with comparable accuracy exhibit poorly separated class clusters or misaligned domain features, as indicated by low Silhouette scores or high MMD. Thus, incorporating these complementary metrics provides a more interpreted and reliable evaluation of generalization behavior.

This comprehensive study establishes Graph Transformers, particularly hybrid GT-MPNN architectures, as a promising foundation for robust, real-world graph learning applications with superior OOD generalization capabilities, and demonstrates the value of domain-alignment and class-separation metrics for deeper model evaluation.

\bibliographystyle{unsrtnat}


\appendix
\renewcommand{\thesection}{\Alph{section}}

\section*{\LARGE Appendix}

\section{Experimental details}\label{sec:exp_details}

\begin{table}[ht]
\centering
\small
\begin{tabular}{lllllllllp{1.5cm}}
\toprule
Dataset / Shift & MPNN & GT & \#Layers & \#Params & BSZ & \#Epochs & LR & WD & Scheduler \\
\midrule
SST2-Length & \ac{vGIN} & -- & 3 & 1.7M & 32 & 200 & 1e-3 & 0 & cosine \\
SST2-Length & -- & MHA & 5 & 3.8M & 32 & 200 & 1e-3 & 0 & step 150 \\
SST2-Length & GIN & MHA & 3 & 2.9M & 32 & 200 & 1e-3 & 0 & cosine \\
CMNIST-Color & \ac{vGIN} & -- & 7 & 2.7M & 128 & 500 & 1e-3 & 1e-4 & step 300 \\
CMNIST-Color & -- & MHA & 4 & 2.9M & 128 & 500 & 1e-3 & 1e-4 & step 300 \\
CMNIST-Color & GIN & MHA & 3 & 2.7M & 128 & 500 & 1e-3 & 1e-4 & step 300 \\
Motif-Size & GIN & -- & 4 & 1.2M & 32 & 200 & 1e-3 & 0 & none \\
Motif-Size & -- & MHA & 5 & 3.6M & 32 & 200 & 1e-3 & 0 & step 150 \\
Motif-Size & GIN & MHA & 4 & 3.6M & 32 & 200 & 1e-3 & 0 & step 150 \\
Motif-Basis & GIN & -- & 4 & 1.2M & 32 & 200 & 1e-3 & 0 & none \\
Motif-Basis & -- & MHA & 5 & 3.6M & 32 & 200 & 1e-3 & 0 & step 150 \\
Motif-Basis & GIN & MHA & 4 & 3.6M & 32 & 200 & 1e-3 & 0 & step 150 \\
HIV-Size & \ac{vGIN} & -- & 3 & 1.5M & 32 & 200 & 1e-3 & 0 & cosine \\
HIV-Size & -- & MHA & 2 & 1.5M & 32 & 200 & 1e-3 & 0 & cosine \\
HIV-Size & GIN & MHA & 3 & 1.8M & 32 & 200 & 1e-3 & 0 & cosine \\
HIV-Scaffold & \ac{vGIN} & -- & 3 & 1.5M & 32 & 200 & 1e-3 & 1e-4 & step 150 \\
HIV-Scaffold & -- & MHA & 2 & 1.5M & 32 & 200 & 1e-3 & 1e-4 & step 150 \\
HIV-Scaffold & GIN & MHA & 2 & 1.8M & 32 & 200 & 1e-3 & 1e-4 & step 150 \\
\bottomrule
\end{tabular}
\caption{Model architectures and training hyperparameters across GOOD datasets and distribution shifts}
\label{tab:hyperparams}
\end{table}

We perform graph-level prediction experiments on three datasets, across six covariate shift splits, using seven DG algorithms and three backbone methods, as detailed in Table \ref{tab:hyperparams}. Each group of three consecutive rows corresponds to the evaluated backbones: vGIN/GIN for the MPNN, MHA for the GT, and GPS as the hybrid. Our implementation builds upon PyTorch Geometric \cite{fey2019fast} and the GOOD benchmark framework \cite{gui2022good}. To ensure a fair comparison, we initially matched the number of parameters across backbones for each task. However, when varying the number of layers, we observed that in some cases smaller models achieved higher accuracy, leading us to select model sizes accordingly.

For experiments on CMNIST and HIV-Scaffold, we use the AdamW optimizer with 1e-4 weight decay, which led to better performance. Across all experiments, we apply a dropout rate of 0.5, ReLU activation, and mean global pooling. The hidden layer size is set to 300. Batch size (BSZ), maximum number of training epochs, and learning rates (LR) are listed in Table \ref{tab:hyperparams}. Two schedulers were used: cosine-annealing with 10-epoch warmup and cosine decay to 10\% LR, and single-step, which decays LR by 0.1 at specified milestone. As part of a preliminary analysis, we found that five-dimensional \ac{RW} embeddings offer greater robustness than the ten-dimensional version.

All models are trained to convergence, and each setup is run four times with different random seeds. Reported results are averaged across these runs. Experiments are generally conducted on a single NVIDIA RTX A5000 or A6000 GPU. All code necessary to reproduce the results presented in this paper will be made publicly available.

For each DG algorithm, one or two algorithm-specific hyperparameters were tuned following the procedure outlined in \cite{gui2022good}. For IRM and Deep CORAL, the penalty loss weight was tuned. For VREx, the weight of the loss variance penalty was tuned. In GroupDRO, the step size was tuned. For DANN, the domain classification penalty weight was tuned. Mixup uses a Beta distribution to randomize the lambda value for mixing two instances, and the distribution's alpha parameter was tuned.

\section{In-distribution experimental results discussion}\label{sec:additional_exp}

\begin{figure}[h]
    \centering
    \includegraphics[width=0.9\linewidth]{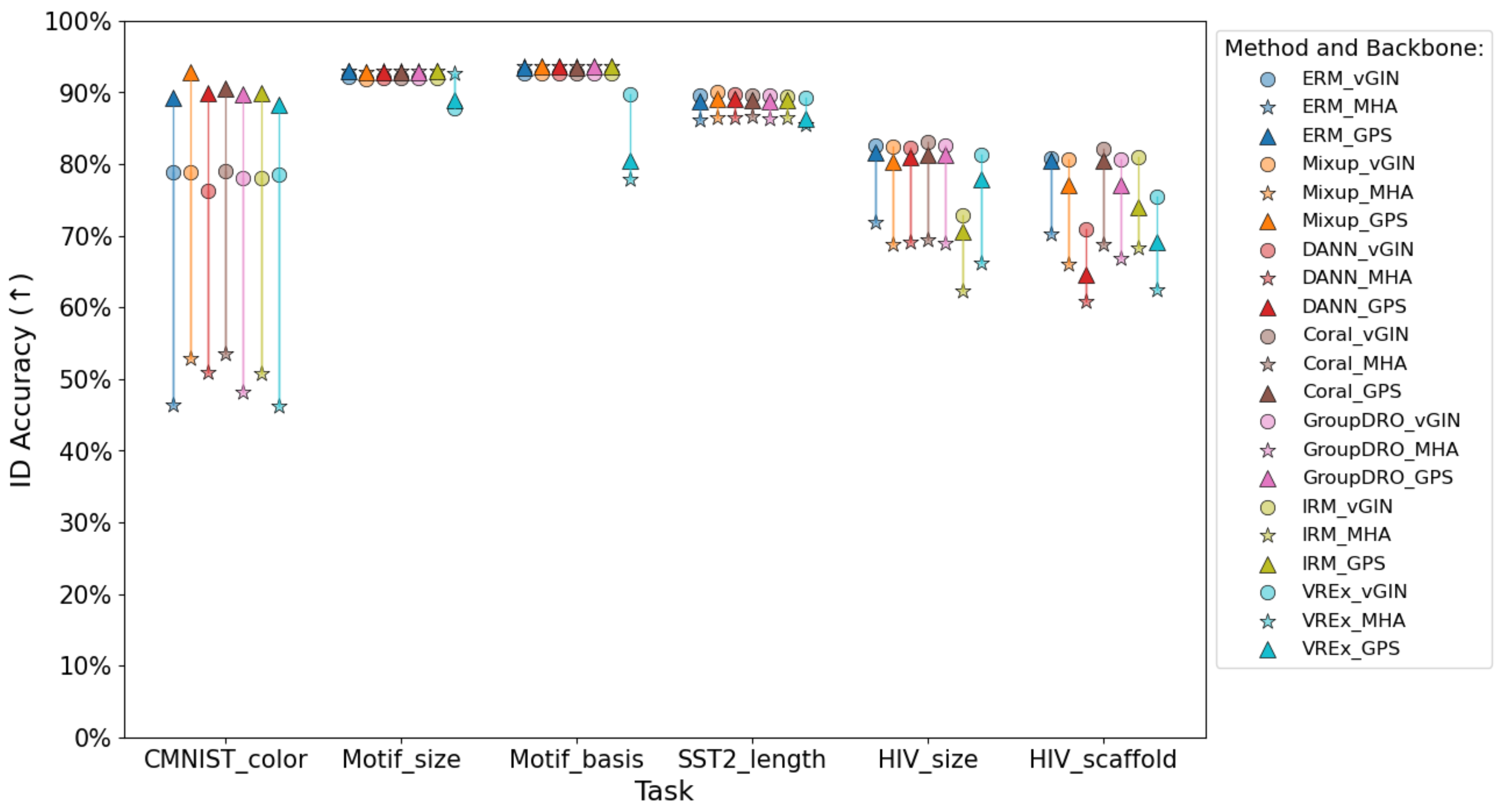}
    \caption{Visualization of ID accuracy results reported in Table \ref{tab:ood_results}. Each column corresponds to a different benchmark, and each color indicates a distinct DG algorithm. Marker shapes represent the backbone: circles for vGIN, stars for MHA, and triangles for GPS.
    }
    \label{fig:id_acc}
\end{figure}

Figure \ref{fig:id_acc} provides a comprehensive visualization of ID test accuracy results across the six benchmark datasets. This analysis complements our OOD findings by validating that high OOD performance does not come at the expense of staying competitive, or even superior, under standard ID conditions. These results also appear in Table \ref{tab:ood_results}. On CMNIST, GPS improves the vGIN results significantly across all DG algorithms, while MHA notably decreases ID accuracy. On Motif tasks, backbones overall perform comparably. On SST2, vGIN slightly outperforms GPS, while MHA shows lower performance.
On the HIV tasks, vGIN generally achieves the best performance, although GPS is comparable in certain cases.
These findings are crucial for practical applications where both ID and OOD samples are encountered, as they show that the superior OOD generalization capabilities of GPS hybrid do not come at the expense of standard ID performance.

\section{Prior work on in-distribution performance of GTs compared to MPNNs}\label{sec:gt_mpnn_id_summary}

To provide additional context, Table \ref{tab:gt_mpnn_id_summary} summarizes key findings from prior research comparing GTs and MPNNs in ID settings. The table highlights differences in their performance across various graph types, architectural advantages such as virtual node augmentation and positional encodings, and identifies gaps in robustness studies under distribution shifts. The corresponding takeaways guided the design choices and evaluation strategies applied in our work.

\begin{table*}[h]
\centering
\small
\begin{tabularx}{\textwidth}{lX X}
\toprule
\textbf{Aspect} & \textbf{Key Insight} & \textbf{Takeaway} \\
\midrule
Heterophilous graphs' performance &
GTs consistently outperform MPNNs in a large-scale empirical study covering diverse datasets that vary in task type, graph size, and sparsity \cite{tang2025opengt}. &
GTs are preferable for tasks in which neighboring nodes are not necessarily similar. \\
Homophilous graphs' performance &
GTs and MPNNs perform comparably \cite{tang2025opengt}. &
GTs remain a safe choice due to their flexible receptive fields. \\
Virtual node augmentation &
Virtual nodes significantly improve MPNN performance and theoretically approximate GT attention under certain conditions \cite{cai2023connection, rosenbluth2024distinguished, nguyen2024alignment}. &
In this work, we incorporated virtual node augmentation to strengthen our MPNN baseline. \\
Architectural design &
A comprehensive empirical study across diverse GT architectures and positional encodings shows that GraphGPS with RW-based encodings achieves SOTA results, highlighting the effectiveness of hybrid architectures and expressive positional encodings \cite{tang2025opengt}. &
This strong empirical evidence motivates our choice to use GPS-RW in this work. \\
Graph reasoning tasks (ID) &
Transformer-based models have been empirically and theoretically shown to excel at graph reasoning, often matching or surpassing \ac{MPNNs} in \ac{ID} scenarios \cite{sanford2024understanding}. &
These results motivate our hypothesis that \ac{GT} may also generalize better than \ac{MPNNs} under distribution shifts. \\
OOD generalization studies &
MPNN robustness to structural and feature distribution shifts has been extensively explored \cite{guo2024investigating, lu2024graph}. &
In contrast, the robustness of GT to such distribution shifts remains under-investigated. \\
\bottomrule
\end{tabularx}
\caption{Summary of key insights from prior work comparing GTs and MPNNs in ID settings, alongside takeaways applied in this study.}
\label{tab:gt_mpnn_id_summary}
\end{table*}



\end{document}